\documentclass[lettersize,journal]{IEEEtran}
\usepackage{amsmath,amsfonts}
\usepackage{algorithmic}
\usepackage{array}
\usepackage[caption=false,font=normalsize,labelfont=sf,textfont=sf]{subfig}
\usepackage{textcomp}
\usepackage{stfloats}
\usepackage{url}
\usepackage{cite}

\usepackage{verbatim}
\usepackage{graphicx}
\def\BibTeX{{\rm B\kern-.05em{\sc i\kern-.025em b}\kern-.08em
    T\kern-.1667em\lower.7ex\hbox{E}\kern-.125emX}}
\usepackage{balance}

\begin{document}

\author{Jacob Gildenblat, Jens Pahnke

\thanks{\textit{J. Gildenblat  and J. Pahnke are with the Translational Neurodegeneration Research and Neuropathology Lab, Department of Clinical Medicine, Medical Faculty, University of Oslo (UiO) and Section of Neuropathology Research, Department of Pathology, Clinics for Laboratory Medicine, Oslo University Hospital (OUS), Norway. Web: www.pahnkelab.eu, e-mails: jacob.gildenblat@gmail.com; jens.pahnke@gmail.com.}}
\thanks{\textit{J.P. is also with the Institute of Nutritional Medicine, University of Lübeck (UzL) and University Medical Center Schleswig-Holstein (UKSH), Germany, the Department of Neuromedicine and Neuroscience, The Faculty of Medicine and Life Sciences, University of Latvia (LU), and the Department of Neurobiology, School of Neurobiology, Biochemistry and Biophysics, The Georg S. Wise Faculty of Life Sciences, Tel Aviv University (TAU), Israel.}}
}

\title{Preserving clusters and correlations: a dimensionality reduction method for exceptionally high global structure preservation}

\maketitle

\begin{abstract}
    We present \textit{Preserving Clusters and Correlations} (PCC), a novel dimensionality reduction (DR) method that achieves state-of-the-art global structure (GS) preservation while maintaining competitive local structure (LS) preservation. It optimizes two objectives: a GS preservation objective that preserves an approximation of Pearson and Spearman correlations between high- and low-dimensional distances, and an LS preservation objective that ensures clusters in the high-dimensional data are separable in the low-dimensional data. PCC has a state-of-the-art ability to preserve the GS while having competitive LS preservation. In addition, we show the correlation objective can be combined with UMAP to significantly improve its GS preservation with minimal degradation of the LS. We quantitatively benchmark PCC against existing methods and demonstrate its utility in medical imaging, and show PCC is a competitive DR technique that demonstrates superior GS preservation in our benchmarks.
\end{abstract}


\section{Introduction}

\IEEEPARstart{D}{imensionality} reduction (DR) methods are widely used in data science, both as a pre-processing technique for machine learning and to visualize data by transforming it into 2 or 3 dimensions. These methods can broadly be categorized into methods that focus on preserving the GS of the high-dimensional data, and those that focus on preserving the LS. PCA \cite{pearson1901} and MDS \cite{torgerson1952} are examples of the former, and t-SNE \cite{maaten2008visualizing} or Isomap \cite{tenenbaum2000global} are examples of the latter.
UMAP is a widely adopted DR method, with increased scalability and GS preservation compared to t-SNE \cite{mcinnes2018umap}.

Despite UMAP's widespread adoption, particularly in life sciences, several studies have highlighted its limitations, including poor GS preservation and sensitivity to initialization. \cite{kobak2021initialization}. Low GS preservation means that the distances between point clusters, and relationships between several points in general are not ensured to be meaningful. Even within clusters of points, local distances within clusters may not reflect true high-dimensional relationships.

Modern DR methods like UMAP, with high LS but low GS preservation, primarily preserve local clusters but do not reliably maintain inter-cluster relationships. This is still highly useful for data exploration and identifying clusters in the data. However, these methods may be misleading when it is required to distinguish between close points in the clusters, analyze global trends in the data, or have a reliable pre-processing step.

To address this, we propose a simple \textit{global correlation loss objective} that excels in preserving GS. We sample data reference points and then for each point in the data, we measure the distances from these reference points. The correlation objective demands a high correlation between these distances in the high-dimensional data and in the learned low-dimensional representation. By enforcing correlation between high- and low-dimensional distances to reference points, PCC preserves the relative positioning of data points, ensuring a faithful low-dimensional representation.

For correlation we consider the Pearson \cite{pearson1895correlation} correlation, and a differential approximation of the Spearman rank correlation \cite{spearman1904general} proposed in \cite{blondel2020fast}. This objective achieves high GS preservation, exceeding all current methods by a very large margin. Notably, it also considerably improves upon PCA, a classical method often considered the gold standard for GS preservation.

Unlike graph-based methods like UMAP \cite{mcinnes2018umap}, t-SNE \cite{maaten2008visualizing} or PaCMAP \cite{JMLR:v22:20-1061}, PCC is conceptually simple, and straightforward to implement. PCC can be connected to existing DR methods to improve GS preservation.

Our main contributions are:

1. A global correlation preservation objective based on sampling reference points and for each data point maximizing the high/low dimensional distance correlations between it and all the reference points.

2. We propose a simple method for preserving LS, requiring processing all points independently without constructing a neighbor graph, based on preserving clustering observability. We use a simple LS objective that preserves cluster separability in the low-dimensional representation, by joint learning of both the transformation and a linear classifier that classifies which clusters transformed points belong to.

3. We combine the correlation objective with UMAP in two different ways. In the first, we optimize for the UMAP objective enhanced with the correlation objective and show that it significantly improves GS with minimal LS degradation. In the second, we consider enhancing precomputed UMAP representations by running a few iterations with the correlation loss.

\section{Previous work}
PCA \cite{pearson1895correlation} is a linear dimensionality reduction technique that finds an orthonormal linear projection of the high-dimensional data that maximizes the variance. Distances between points that are along the principle components hyper-planes are fully preserved, while for other points their proximity to the principle components determines their distance preservation. Therefore PCA is a useful method for high GS preservation and is often considered a gold standard method for this, although if the selected number of components does not cover the variation in the data, distortions are expected.

UMAP \cite{mcinnes2018umap} is a widely used non-linear dimensionality reduction technique that constructs a high-dimensional graph of data relationships and embeds it into a lower-dimensional space using a fuzzy topological framework. The focus of UMAP and similar methods is on preserving the local neighborhood of points through the neighbor graph and thus on the LS. UMAP can preserve some GS through an initialization that preserves GS, for example by initializing with PCA.

PaCMAP \cite{JMLR:v22:20-1061} is a method that follows up on UMAP to improve the GS preservation explicitly. It achieves this by constructing a graph with three types of edges: near pairs, mid-range pairs, and far pairs.

Finally, Parametric-UMAP \cite{sainburg2021parametric} trains a parametric model with the UMAP objective with mini-batch gradient descent. They propose a GS preservation loss term by measuring distances between points in the batch and maximizing the Pearson correlation between the distances in the high and low dimensional spaces. This requires re-computing distances between all pairs in a batch, potentially limiting the batch size. We expand upon this idea for a non-parametric transformation, without mini batches, considering all data points at once.

\section{Method}
\subsection{The Correlation objective for global structure preservation}

    \begin{figure}[h!]
        \centering
        \includegraphics[width=3.4in]{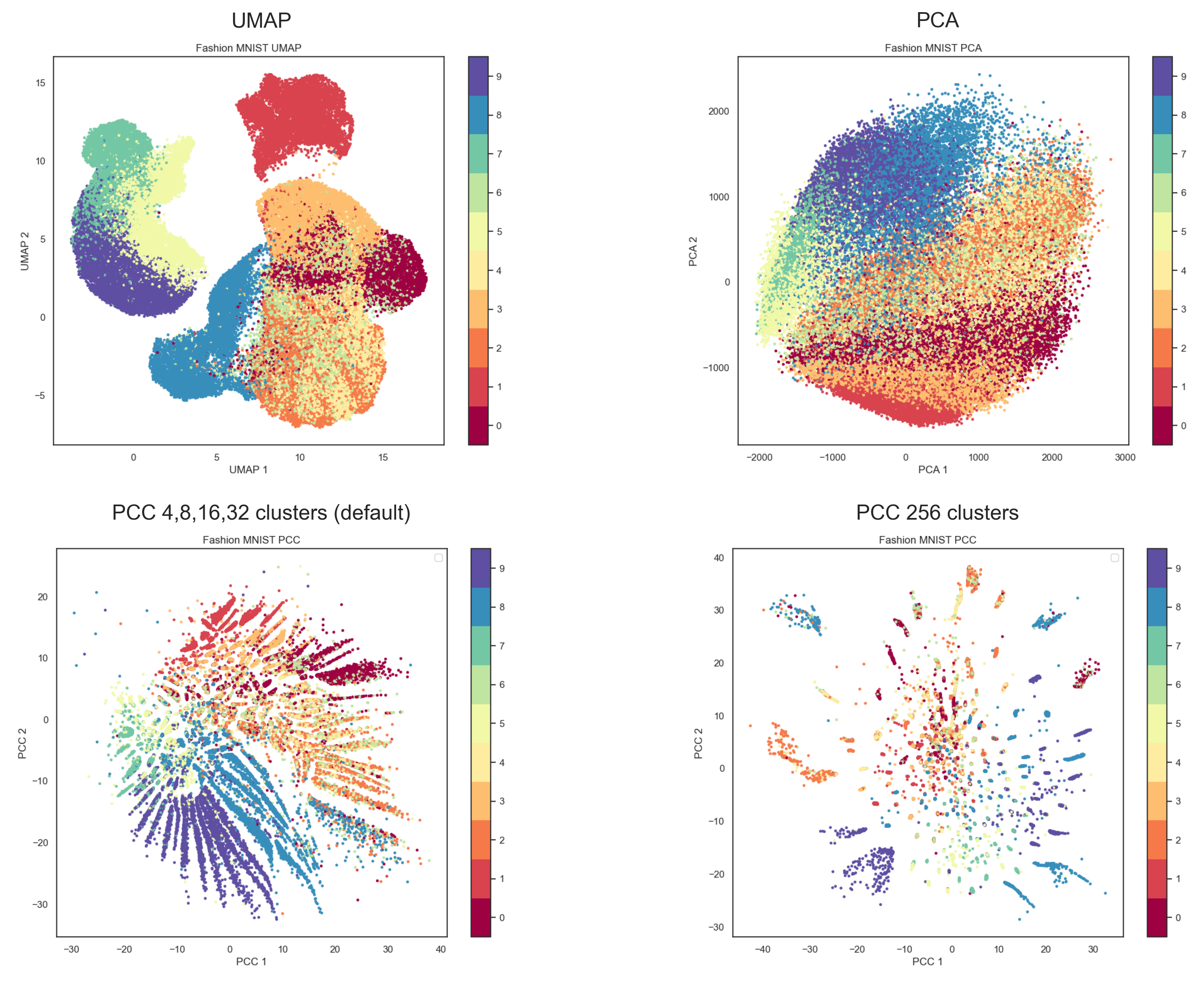}
        \caption{Results on Fashion MNIST. In PCC, unlike UMAP, distances between different points are meaningful since GS is preserved. Unlike PCA, clusters are separated because of the higher GS. By using a higher cluster choice like 256, we get isolated groups of points belonging to those clusters.}
        \label{fig:example}
    \end{figure}

The global correlation metric proposed in \cite{geng2005supervised} measures the correlation of pairs of points in the high-dimensional data, and the transformed data, to evaluate how well the GS is preserved. Motivated by this metric, we aim to approximate it in a differentiable way and optimize for it directly.

Consider a group of points $\{ \mathbf{x}_i \}_{i=1}^{N}$ that we want to transform to a lower dimensionality $y_i$. We sample a subset of K indices of points in $x$: 
$$I = \{ i_1, i_2, \dots, i_K \}, \quad i_1, i_2, \dots, i_K \in \{1, 2, \dots, N\}$$

$I_j$ is then the index of the j'th sampled data point in $x$. 

For a distance function D, e.g. the Euclidean distance, we measure the distances between each data point and all the reference points in the high dimensional data $x$ and the transformation $y$: 

$$d^x_{ij} = D(x_i, x_{I_j}) = ||x_i - x_{I_j}||_2$$
$$d^y_{ij} = D(y_i, y_{I_j})= ||y_i - y_{I_j}||_2$$

The correlation loss objective is to maximize the correlation between $d^x$ and $d^y$. For the Pearson correlation this is

$$L_{\text{Pearson}} = -\frac{\text{Cov}(d^x, d^y)}{\sqrt{\text{Var}(d^x)} \sqrt{\text{Var}(d^y)}}$$
$$ = -\frac{\sum_{i=1}^{N}\sum_{j=1}^{K}(d^x_{ij} - \bar{d}^x)(d^y_{ij} - \bar{d}^y)}
{\sqrt{\sum_{i=1}^{N}\sum_{j=1}^{K}(d^x_{ij} - \bar{d}^x)^2} \sqrt{\sum_{i=1}^{N}\sum_{j=1}^{K}(d^y_{ij} - \bar{d}^y)^2}}
$$

We also consider the Spearman rank correlation (and the average of both correlations).
Although the Pearson correlation implementation is differentiable and can be maximized with gradient descent, the Spearman rank correlation is not, because ranks are not differentiable. We use a differentiable approximation of the ranks, the soft ranking proposed in \cite{blondel2020fast} and perform the correlation on that.
For Spearman rank correlation, for each point we compute the soft rank approximations of the distances from the point to the k reference points and measure the correlation of these approximated ranks with the ranks of the same distances in the transformed lower-dimensional data.

$r^x_{ij} \in [1, K]$ are the ranks of the data point $i$ from reference points $j$, ranked against the K reference points.
$r^y_{ij}$ is the approximated soft rank of the lower dimensional data point $i$ from reference point $j$, ranked against the K reference points.

$$L_{\text{spearman}} = -\frac{\text{Cov}(r^x, r^y)}{\sqrt{\text{Var}(r^x)} \sqrt{\text{Var}(r^y)}}
$$

Finally, for the correlation loss, we use the average of the Pearson and Spearman correlation losses.
$$L_{\text{correlation}} = 0.5L_{\text{Pearson}} + 0.5L_{\text{Spearman}}$$

\subsection{The clustering observability objective for local structure preservation}

DR methods with good LS behavior are often used to visualize or detect clusters in the data. In 2D scatter plots, DR points are often colorized by the clusters they belong to. We propose reversing this, directly clustering the original data, and then learning a reduced dimensionality embedding where it is possible to predict for every point which cluster (or clusters) it belongs to.

We hypothesize that given a good clustering model, the close neighborhood of a point in the high-dimensional data is more likely to belong to the same clusters. Therefore, if the LS is preserved in the transformed data, neighboring points should in most cases belong to the same clusters, and it should be possible to predict what cluster a transformed point belongs to.
On the other hand, if it is not possible to predict which cluster a transformed data point belongs to, it means that points belonging to the same cluster are not grouped.

Motivated by this, we define a simple LS objective, by predicting the clusters. Given a clustering model that assigns each data point to one of k clusters, we learn a linear classifier A on top of the (also learned) transformed visualization that predicts the assigned cluster. For the linear classifier to be able to predict the clusters, the clusters have to be separable in the learned low-dimensional transformation. Thus, the joint learning of the cluster classifier and the transformation encourages separating or grouping data points according to their cluster.

$$\mathcal{L_\text{cluster}} = -\sum_{i} y_i \log((A \cdot e)_i)$$

The weights of this classifier are optimized jointly with the low-dimensional transformation.

In practice, we use several clustering models for different numbers of clusters with a multi-task loss function:
$$\mathcal{L_\text{cluster}} = -\frac{1}{M}\sum_{m=1}^{M} \sum_{i} y_i \log((A^{(m)} \cdot e)_i)$$

\subsection{Combining the local and global objectives}
Deviating from UMAP/t-SNE, we use a random normal initialization and do not rely on initialization from PCA.

The loss is then  $$L_{PCC} = L_\text{cluster} + \beta \cdot L_\text{correlation}$$

For the cluster observability objective when applied, we use a multi-task objective, predicting all clusters, with k = 4, 8, 16, 32, or 64.

Examples for the Fashion-MNIST \cite{tensorflow_fashion_mnist} dataset are shown in \ref{fig:example}. PCC tends to create lines with points belonging to the same clusters created by the linear cluster assignment classifier. Different cluster choices affect the visualization output, with more clusters causing more isolated regions. 

\section{Experiments}

\subsection{Benchmarking}
We benchmarked PCC against several modern DR methods: UMAP \cite{mcinnes2018umap}, t-SNE \cite{maaten2008visualizing}, TriMap \cite{amid2019trimap}, PaCMAP \cite{JMLR:v22:20-1061} and PHATE \cite{moon2017phate}. For LS evaluation, we use the Trustworthiness and Continuity metrics \cite{lee2007nonlinear} and the Mean Relative Rank Error metrics \cite{lee2009quality}. For GS evaluation, we use Pearson and Spearman correlations as proposed in \cite{geng2005supervised}. For computing all metrics, we use the recent Zadu python library \cite{jeon23vis}.

We evaluated 9 datasets covering different use cases: the MNIST \cite{mnist_dataset} dataset and Fashion-MNIST \cite{tensorflow_fashion_mnist} dataset often used as a more challenging alternative to MNIST. For samples of life sciences datasets, we used Macosko single-cell dataset \cite{macosko2015}. For examples of deep learning embeddings, we used ResNet50 \cite{he2016deep} embeddings of CIFAR \cite{krizhevsky2009learning}, CIFAR100 \cite{krizhevsky2009learning}, and the miniImageNET \cite{ren2018meta} datasets. Finally, for simple synthetic datasets employed to diagnose issues with DR methods, we used the mammoth\cite{understandingumap2020} and Swiss roll \cite{tenenbaum2000global} datasets.

\subsection{PCUMAP - Combining the global loss for Preserving Correlations with UMAP}
We tested if we can improve the low GS in UMAP using the global correlation loss. We used a weight of 0.001 for the UMAP loss and added it to the PC loss. We used the TorchDR python package \cite{torchdr}. To help UMAP converge, we first run 10 iterations of the UMAP training before adding the global correlation loss.

$$L_{\text{PCUMAP}} = L_{\text{UMAP}} + \beta \cdot L_{corr}$$

\begin{figure}[h!]
        \centering
        \includegraphics[width=3.4in]{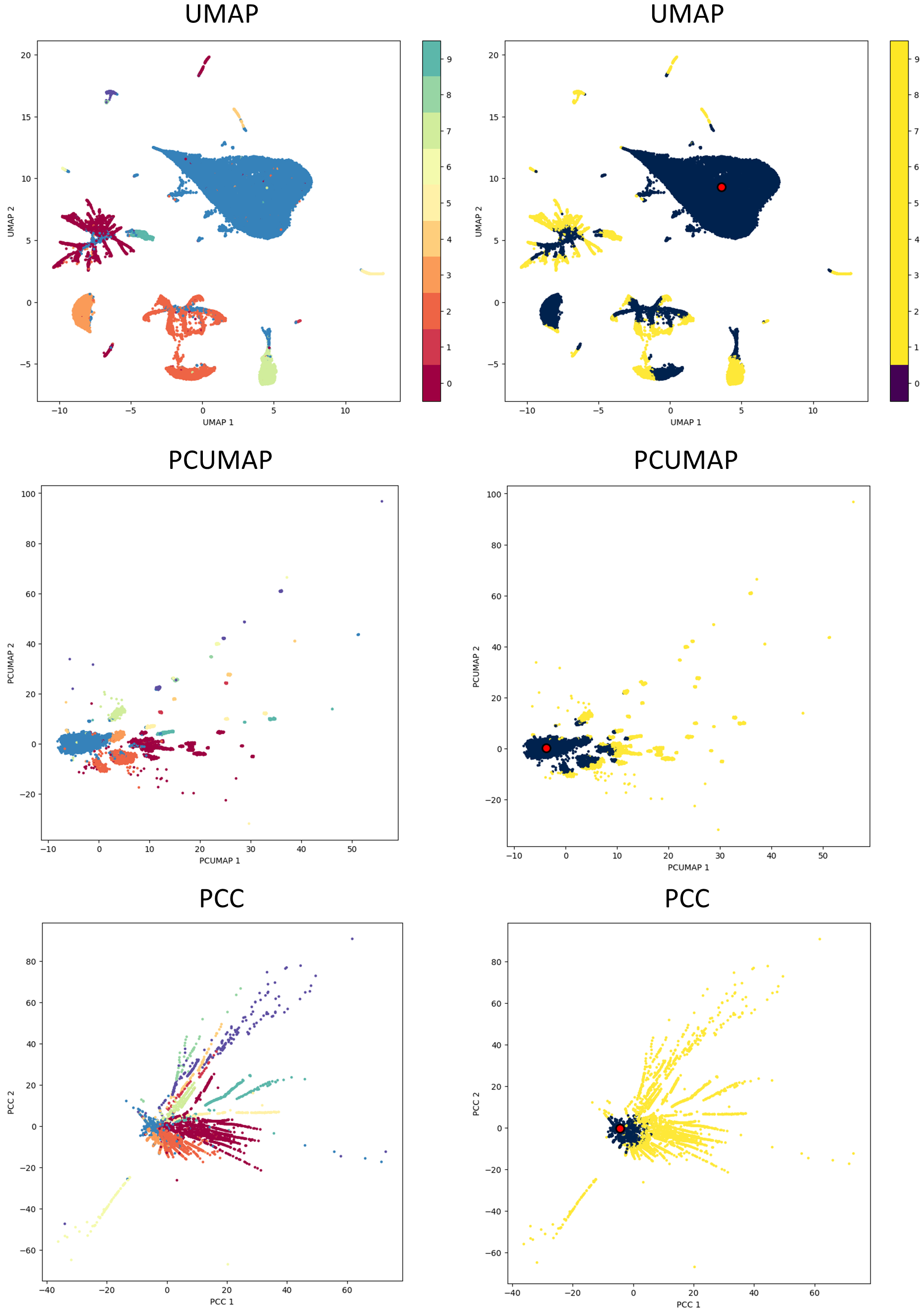}
        \caption{Comparing UMAP (an existing method) and PCUMAP and PCC (our proposed methods) on the Macosko single cell dataset \cite{macosko2015}. Upper row: The transformed data is colored by labels. Bottom row: colored according to distances from a selected point, in the high dimensional data. In UMAP the points in the low-dimensional data do not preserve the original distance: many points far away are close points in the high-dimensional data. In PCUMAP and PCC this is solved.}
        \label{fig:PCUMAP}
    \end{figure}

Figure \ref{fig:PCUMAP} presents the application of PCUMAP to the Macosko single-cell dataset \cite{macosko2015}, alongside PCC and UMAP. The bottom row visualizes the transformed data, with points colored based on their (thresholded) distances from a selected reference point, marked in red. In UMAP, both near (yellow) and distant (blue) points appear interspersed around the selected point, highlighting its poor global structure preservation. As a result, the relative distances between points are not reliable indicators of their true high-dimensional relationships. In contrast, PCUMAP and PCC exhibit a more structured separation, where nearby and distant points are clearly delineated, better reflecting the high-dimensional distance relationships.

\subsection{Initializing  from UMAP and then running the global correlation objective}
Here, we tested if we can add GS to an existing UMAP transformation. We initialized from UMAP and ran 3 iterations with the correlation loss with an additional mean squared error loss term that makes sure the result does not deviate too much from the initialization.

$$L_\text{UMAP init + PC} = L_\text{corr} + \lambda||e - e_\text{UMAP}||^2$$

$\lambda$ controls how close we want to keep to the initial embedding. We use $\lambda=1$.

\subsection{Comparing UMAP and PCC visualizations of biological data}
To assess the practical performance of PCC for the visualization of life science / biological data, we used mass spectrometry imaging (MSI) lipidomics data from an Alzheimer's disease mouse model that was generated by us using a timsTOF fleX™ mass spectrometer in MALDI mode (Bruker Daltonics, Bremen, Germany).
MSI is a challenging test case for DR methods since these images contain a large amount of detail that will not be revealed if the DR methods merely separate the data into clusters. 

\section{Results}

\subsection{Benchmarking}
\begin{figure}[h!]
        \centering
        \includegraphics[width=3.4in]{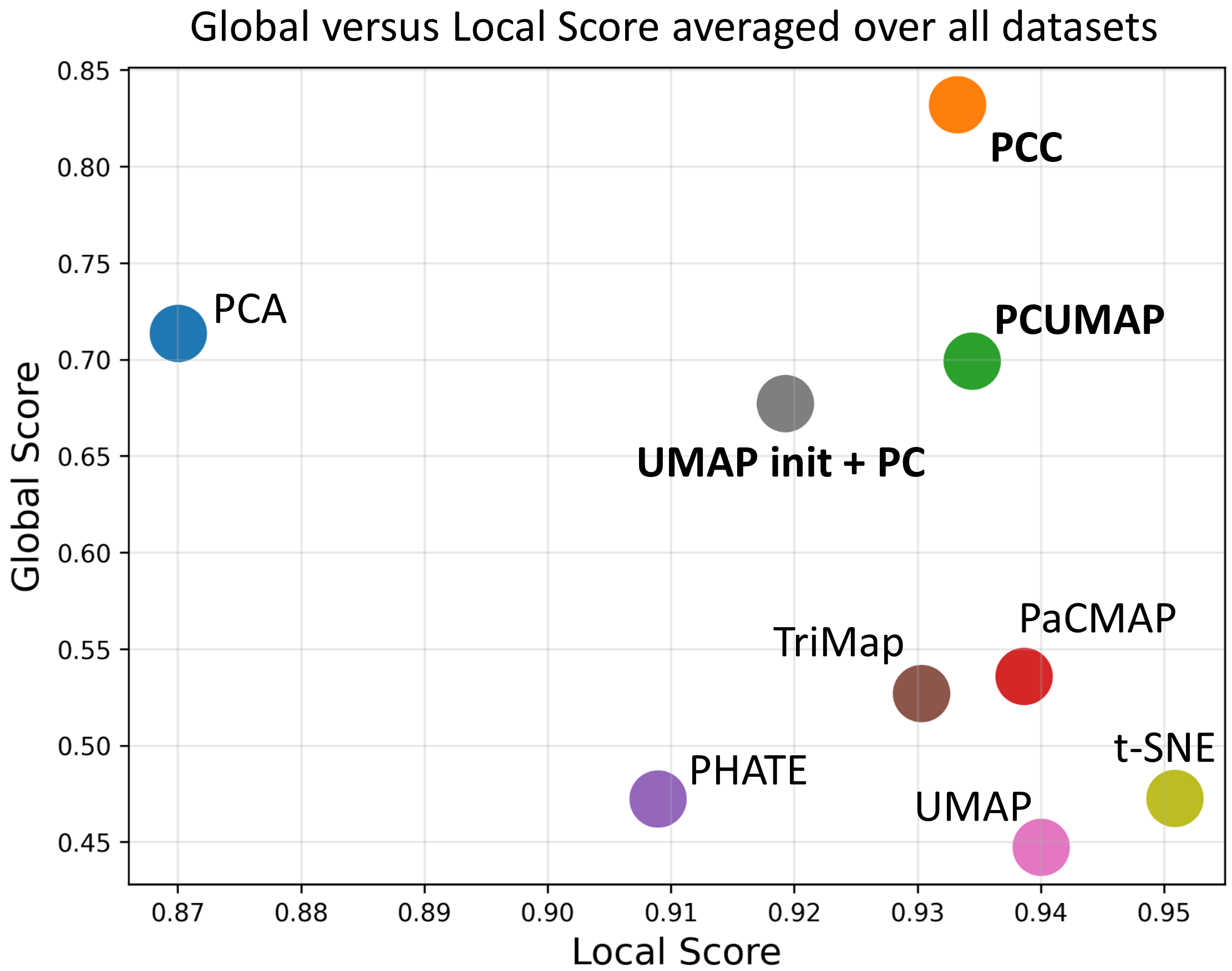}
        \caption{Plotting the average performance of GS metrics against local structure metrics on 9 datasets. Our proposed methods: \textit{\textbf{PCC, UMAP init+PC, PCUMAP.}}
        PCC improves global structure preservation over all other tested methods by a large margin (average of 0.83, while PCA gets 0.71 and UMAP 0.44), while being competitive in local structure preservation with graph methods that specialize in local structure (e.g, PCC gets 0.933 and UMAP 0.94). Among the modern graph methods, PaCMAP performs the best and slightly improves the global structure compared to UMAP. However, there is still room for improvement in global structure preservation, which we show is possible.}
        \label{fig:results_scoring}
    \end{figure}

Figure \ref{fig:results_scoring} shows the benchmarking results averaged over the 9 tested datasets. As a single LS metric, we take the average of the 4 LS metrics. Similarly, for a GS metric, we used the average of the Pearson and Spearman correlation metrics.

PCA is a gold standard method for high GS preservation, but it is lower in LS preservation. Amongern methods, PaCMAP and TriMap improve over t-SNE/UMAP, with the PaCMAP method indeed performing the best, while PHATE performs the worst. PCC improves the GS preservation over all other tested methods by a large margin and is still a competitive method for LS preservation despite the clustering objective simplicity, making it a competitive method when the global data structure should be preserved. The full results are given in Table \ref{results-table}.

Combined optimization of UMAP and the global loss (PCUMAP) is able to achieve GS close the PCA, and a high improvement compared to UMAP alone, with a small degradation in the LS.

Initialization from UMAP and 3 PC iterations (UMAP unit + PC) is also able to improve the GS substantially, however, results in losing LS.

\subsection{Comparison of UMAP and PCC using MSI visualization}
\begin{figure}[h!]
        \centering
        \includegraphics[width=3.4in]{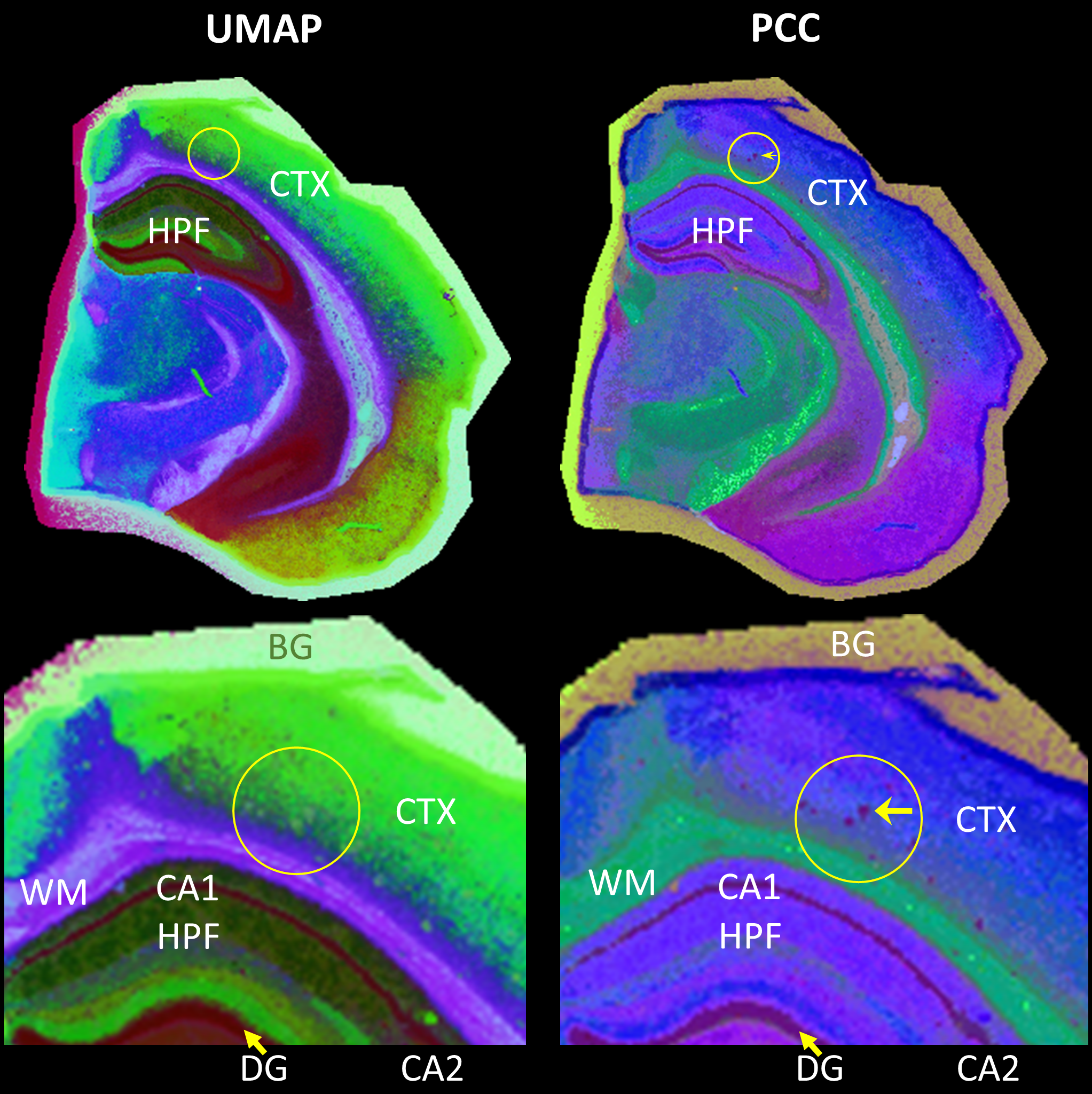}
        \caption{Comparing UMAP and PCC using visualizations of lipidomics MSI data of mouse brain. Both methods reduce the high-dimensional image data to 3 dimensions which are then normalized and colored as RGB images. PCC reveals numerous pathological changes, so-called $\beta$-amyloid plaques (circle and arrow, brown dots) in the isocortex (CTX) of the Alzheimer's disease mouse model, but also enables visualization of normal structures, e.g. the neuronal band of the dentate gyrus (DG, arrow, band) in the hippocampus formation (HPF). Both structures are in-detectable using UMAP visualization (left images). Legend: CTX - isocortex of the cerebrum, HPF - hippocampus formation, DG - dentate gyrus, CA1 and CA2 - cornu ammonis neurons, area 1 and 2, WM - white matter, BG - background.}
        \label{fig:results_brain}
    \end{figure}

In addition to the quantitative improvement on the MSI mouse model dataset in the benchmark, Figure \ref{fig:results_brain} shows a comparison of UMAP and PCC to visualize lipidomics MSI  data of a mouse brain hemisphere. The preservation of the global data structure improves significantly the visualization of pathological tissue changes (e.g., $\beta$-amyloid plaques as found in this Alzheimer's disease mouse model) but also allows the detection of normal anatomical structures (e.g., the neuronal band of the dentate gyrus).

\section{Discussion}
This paper presents PCC, a dimensionality reduction method that focuses on improving GS preservation. Compared to previous methods, PCC achieves the highest global structure preservation by a large margin. This is achieved by maximizing the correlations of the distances of all points from a set of reference points in the high-dimensional and low-dimensional data. This deviates from modern DR methods that focus on neighbor graph construction and achieving high LS preservation. We show that follow-up methods on UMAP like TriMap and PaCMAP meant to solve the GS preservation problems of UMAP only marginally improve GS preservation as compared to our new PCC method. We also show that we can plug the global correlation loss into methods like non-parametric UMAP to improve their GS preservation. For local structure preservation, we proposed a simple multi-task cluster observability objective that is able to achieve competitive local structure preservation much better than PCA and even PHATE. However, it is still closely behind graph-based methods like UMAP. Follow-up work on PCC could include improving the clustering objective, combining the correlation objective with objectives of methods like UMAP, and adaptive sampling strategies for the reference points. Overall, when GS preservation is needed, PCC offers a DR option that is significantly improved over previously used methods.

\section{Appendix}
\subsection{Acknowledgements}
We thank Dr Jorunn Stamnes and Thomas Brüning for tissue preparation and performing MSI measurements. Mass spectrometry-based analyses of mouse brains were performed at the Proteomics Core Facility at the University of Oslo/Oslo University Hospital, which is supported by the Core Facilities Program of the South-Eastern Norway Regional Health Authority (HSØ) and NAPI (www.napi.uio.no, NFR, Norway; 295910).

\subsection{Funding Information}
J.P. received funding from Norges forskningsråd [NFR, Norway; 327571 (PETABC), 295910 (NAPI, www.napi.uio.no)], DFG (Germany; 263024513), HelseSØ (Norway; 2022046), and the EIC Pathfinder Open Challenges program (EU commission; 101185769).

\subsection{Data availability}
The mass spectrometry imaging dataset of the mouse brain hemisphere is available at ProteomeXchange PRIDE repository with dataset number PXD056609.

\begin{table}[h!]
    \caption{Benchmarking results of DR methods for visualization on 9 datasets.}
    
    \resizebox{\columnwidth}{!}{
    \begin{tabular}{llrrrrrr}
    \hline
    Dataset & Method & Trustworthiness & Continuity & Mrre-False & Mrre-Missing & Pearson & Spearman \\
    \hline
    fashion\_mnist & UMAP & 0.972 & 0.976 & 0.972 & 0.981 & 0.599 & 0.586 \\
    fashion\_mnist & pca & 0.915 & 0.971 & 0.913 & 0.974 & 0.880 & 0.879 \\
    fashion\_mnist & t-SNE & 0.980 & 0.974 & 0.984 & 0.979 & 0.646 & 0.640 \\
    fashion\_mnist & PaCMAP & 0.968 & 0.976 & 0.968 & 0.981 & 0.629 & 0.621 \\
    fashion\_mnist & TriMap & 0.964 & 0.981 & 0.965 & 0.984 & 0.682 & 0.677 \\
    fashion\_mnist & Phate & 0.942 & 0.974 & 0.945 & 0.979 & 0.641 & 0.651 \\
    mouse\_msi & UMAP & 0.963 & 0.961 & 0.963 & 0.966 & 0.695 & 0.770 \\
    mouse\_msi & pca & 0.921 & 0.968 & 0.923 & 0.970 & 0.988 & 0.985 \\
    mouse\_msi & t-SNE & 0.992 & 0.977 & 0.994 & 0.982 & 0.296 & 0.313 \\
    mouse\_msi & PaCMAP & 0.962 & 0.960 & 0.962 & 0.962 & 0.816 & 0.830 \\
    mouse\_msi & TriMap & 0.938 & 0.914 & 0.940 & 0.879 & 0.100 & 0.791 \\
    mouse\_msi & Phate & 0.967 & 0.983 & 0.969 & 0.986 & 0.874 & 0.872 \\
    mammoth & UMAP & 0.990 & 0.978 & 0.992 & 0.984 & 0.774 & 0.800 \\
    mammoth & pca & 0.967 & 0.992 & 0.959 & 0.991 & 0.992 & 0.991 \\
    mammoth & t-SNE & 0.984 & 0.979 & 0.990 & 0.985 & 0.775 & 0.788 \\
    mammoth & PaCMAP & 0.978 & 0.982 & 0.981 & 0.986 & 0.875 & 0.876 \\
    mammoth & TriMap & 0.972 & 0.989 & 0.973 & 0.990 & 0.962 & 0.965 \\
    mammoth & Phate & 0.941 & 0.968 & 0.953 & 0.976 & 0.236 & 0.287 \\
    swiss\_roll & UMAP & 0.997 & 0.981 & 0.996 & 0.988 & 0.423 & 0.377 \\
    swiss\_roll & pca & 0.876 & 0.977 & 0.891 & 0.977 & 0.847 & 0.845 \\
    swiss\_roll & t-SNE & 0.986 & 0.976 & 0.992 & 0.983 & 0.641 & 0.618 \\
    swiss\_roll & PaCMAP & 0.984 & 0.985 & 0.988 & 0.989 & 0.695 & 0.672 \\
    swiss\_roll & TriMap & 0.962 & 0.988 & 0.971 & 0.989 & 0.824 & 0.813 \\
    swiss\_roll & Phate & 0.884 & 0.977 & 0.889 & 0.982 & 0.409 & 0.367 \\
    macosko & UMAP & 0.921 & 0.960 & 0.933 & 0.969 & 0.592 & 0.769 \\
    macosko & pca & 0.745 & 0.904 & 0.744 & 0.912 & 0.924 & 0.937 \\
    macosko & t-SNE & 0.923 & 0.958 & 0.943 & 0.965 & 0.442 & 0.591 \\
    macosko & PaCMAP & 0.920 & 0.964 & 0.929 & 0.970 & 0.638 & 0.798 \\
    macosko & TriMap & 0.907 & 0.967 & 0.918 & 0.972 & 0.671 & 0.795 \\
    macosko & Phate & 0.829 & 0.923 & 0.838 & 0.937 & 0.724 & 0.859 \\
    mnist & UMAP & 0.942 & 0.933 & 0.945 & 0.951 & 0.315 & 0.283 \\
    mnist & pca & 0.739 & 0.908 & 0.737 & 0.922 & 0.536 & 0.505 \\
    mnist & t-SNE & 0.954 & 0.937 & 0.968 & 0.953 & 0.362 & 0.332 \\
    mnist & PaCMAP & 0.935 & 0.928 & 0.938 & 0.947 & 0.332 & 0.296 \\
    mnist & TriMap & 0.919 & 0.932 & 0.922 & 0.950 & 0.196 & 0.190 \\
    mnist & Phate & 0.855 & 0.935 & 0.860 & 0.951 & 0.313 & 0.287 \\
    cifar & UMAP & 0.900 & 0.920 & 0.904 & 0.932 & 0.402 & 0.409 \\
    cifar & pca & 0.771 & 0.894 & 0.769 & 0.903 & 0.544 & 0.534 \\
    cifar & t-SNE & 0.923 & 0.911 & 0.939 & 0.924 & 0.479 & 0.467 \\
    cifar & PaCMAP & 0.887 & 0.913 & 0.886 & 0.924 & 0.395 & 0.415 \\
    cifar & TriMap & 0.877 & 0.916 & 0.880 & 0.927 & 0.410 & 0.408 \\
    cifar & Phate & 0.836 & 0.905 & 0.840 & 0.919 & 0.412 & 0.417 \\
    cifar100 & UMAP & 0.893 & 0.901 & 0.899 & 0.921 & 0.424 & 0.412 \\
    cifar100 & pca & 0.721 & 0.863 & 0.717 & 0.877 & 0.416 & 0.393 \\
    cifar100 & t-SNE & 0.909 & 0.896 & 0.930 & 0.915 & 0.417 & 0.403 \\
    cifar100 & PaCMAP & 0.881 & 0.885 & 0.884 & 0.906 & 0.261 & 0.266 \\
    cifar100 & TriMap & 0.851 & 0.903 & 0.856 & 0.921 & 0.376 & 0.369 \\
    cifar100 & Phate & 0.815 & 0.901 & 0.820 & 0.919 & 0.475 & 0.468 \\
    imagenetmini & UMAP & 0.852 & 0.883 & 0.885 & 0.918 & 0.141 & 0.131 \\
    imagenetmini & pca & 0.665 & 0.837 & 0.661 & 0.851 & 0.330 & 0.320 \\
    imagenetmini & t-SNE & 0.849 & 0.881 & 0.900 & 0.917 & 0.154 & 0.144 \\
    imagenetmini & PaCMAP & 0.830 & 0.884 & 0.850 & 0.919 & 0.122 & 0.111 \\
    imagenetmini & TriMap & 0.817 & 0.891 & 0.838 & 0.924 & 0.132 & 0.124 \\
    imagenetmini & Phate & 0.758 & 0.878 & 0.775 & 0.912 & 0.107 & 0.107 \\
    fashion\_mnist & UMAP init + PC & 0.954 & 0.976 & 0.954 & 0.979 & 0.800 & 0.794 \\
    mouse\_msi & UMAP init + PC & 0.960 & 0.960 & 0.960 & 0.965 & 0.775 & 0.814 \\
    mammoth & UMAP init + PC & 0.985 & 0.983 & 0.986 & 0.985 & 0.864 & 0.893 \\
    swiss\_roll & UMAP & 0.996 & 0.979 & 0.996 & 0.987 & 0.393 & 0.339 \\
    swiss\_roll & UMAP init + PC & 0.985 & 0.976 & 0.985 & 0.983 & 0.553 & 0.510 \\
    macosko & UMAP init + PC & 0.898 & 0.944 & 0.911 & 0.953 & 0.694 & 0.813 \\
    mnist & UMAP init + PC & 0.889 & 0.937 & 0.890 & 0.951 & 0.519 & 0.498 \\
    cifar & UMAP init + PC & 0.837 & 0.910 & 0.837 & 0.917 & 0.659 & 0.668 \\
    cifar100 & UMAP & 0.892 & 0.901 & 0.899 & 0.921 & 0.421 & 0.410 \\
    cifar100 & UMAP init + PC & 0.778 & 0.881 & 0.783 & 0.891 & 0.715 & 0.728 \\
    imagenetmini & UMAP & 0.847 & 0.883 & 0.883 & 0.918 & 0.141 & 0.130 \\
    imagenetmini & UMAP init + PC & 0.751 & 0.888 & 0.765 & 0.907 & 0.452 & 0.443 \\
    fashion\_mnist & PCC & 0.962 & 0.973 & 0.965 & 0.975 & 0.873 & 0.878 \\
    mouse\_msi & PCC & 0.956 & 0.955 & 0.958 & 0.956 & 0.978 & 0.975 \\
    mammoth & PCC & 0.970 & 0.987 & 0.975 & 0.985 & 0.981 & 0.979 \\
    swiss\_roll & PCC & 0.957 & 0.977 & 0.970 & 0.980 & 0.835 & 0.830 \\
    macosko & PCC & 0.898 & 0.945 & 0.906 & 0.949 & 0.967 & 0.946 \\
    mnist & PCC & 0.878 & 0.929 & 0.885 & 0.936 & 0.726 & 0.759 \\
    cifar & PCC & 0.864 & 0.916 & 0.871 & 0.923 & 0.670 & 0.656 \\
    cifar100 & PCC & 0.828 & 0.893 & 0.837 & 0.902 & 0.650 & 0.611 \\
    imagenetmini & PCUMAP & 0.813 & 0.891 & 0.830 & 0.920 & 0.287 & 0.234 \\
    cifar100 & PCUMAP & 0.857 & 0.908 & 0.859 & 0.924 & 0.524 & 0.508 \\
    cifar & PCUMAP & 0.873 & 0.920 & 0.873 & 0.930 & 0.555 & 0.539 \\
    mnist & PCUMAP & 0.929 & 0.916 & 0.931 & 0.933 & 0.504 & 0.496 \\
    macosko & PCUMAP & 0.910 & 0.962 & 0.923 & 0.966 & 0.888 & 0.894 \\
    swiss\_roll & PCUMAP & 0.921 & 0.984 & 0.930 & 0.986 & 0.864 & 0.858 \\
    mammoth & PCUMAP & 0.969 & 0.992 & 0.966 & 0.990 & 0.989 & 0.988 \\
    fashion\_mnist & PCUMAP & 0.965 & 0.979 & 0.965 & 0.981 & 0.792 & 0.790 \\
    mouse\_msi & PCUMAP & 0.985 & 0.986 & 0.985 & 0.987 & 0.937 & 0.943 \\
    \hline
    \end{tabular}}
    \label{results-table}
\end{table}


\bibliographystyle{IEEEtran}

\end{document}